\documentclass[11pt] {article}

 \usepackage{color}
 \usepackage{amsmath}
\usepackage{amsfonts}
\usepackage{amssymb}
\usepackage{amsthm}
\usepackage{graphicx}
\usepackage{algorithm,algorithmic}
\usepackage{authblk}

 \newcommand{\R} {\mathbb{R}}

\begin{document}    

\title{Option Compatible Reward Inverse Reinforcement Learning  }
\author[a]{Rakhoon Hwang}
\author[b]{Hanjin Lee}
\author[a]{Hyung Ju Hwang\footnote{Correponding address: Department of Mathematics, Pohang University of Science and Technology (POSTECH), 37673, 77, Cheongam-ro, Nam-gu, Pohang, Gyeongsangbuk-do, Republic of Korea\\ E-mail address: hjhwang@postech.ac.kr}}

\affil[a]{Department of Mathematics, Pohang University of Science and Technology (POSTECH), Pohang, Republic of Korea}
\affil[b]{Global leadership school, Handong global university, Pohang, Republic of Korea}
\date{}

\maketitle
\begin{abstract}
Reinforcement learning in complex environments is a challenging problem. In particular, the success of reinforcement learning algorithms depends on a well-designed reward function. Inverse reinforcement learning (IRL) solves the problem of recovering reward functions from expert demonstrations. In this paper, we solve a hierarchical inverse reinforcement learning problem within the options framework, which allows us to utilize intrinsic motivation of the expert demonstrations. A gradient method for parametrized options is used to deduce a defining equation for the Q-feature space, which leads to a reward feature space. Using a second-order optimality condition for option parameters, an optimal reward function is selected. Experimental results in both discrete and continuous domains confirm that our recovered rewards provide a solution to the IRL problem using temporal abstraction, which in turn are effective in accelerating transfer learning tasks. We also show that our method is robust to noises contained in expert demonstrations.\footnote{This paper is under consideration at Pattern Recognition Letters.}
\end{abstract}
 
\normalsize
\section{Introduction}
Reinforcement learning (RL) method seeks an optimal policy for a given reward function in a Markov decision process (MDP). There are several circumstances in which an agent can learn only from an expert demonstration, because it is difficult to prescribe a proper reward function for a given task. Inverse reinforcement learning (IRL) aims to find a reward function that can explain the expert's behavior. When the IRL method is applied to a complex environment, the size of each trajectory of the required demonstration by the expert can be huge. There are also certain complex tasks that must be segmented into a sequence of sub-tasks (e.g., robotics of ubiquitous general-purpose automation (\cite{Konidaris} \cite{Krishnan19}), robotic surgical procedure training (\cite{Fox}, \cite{Krishnan17}), hierarchical human behavior modeling \cite{Solway}, and autonomous driving \cite{Liaw}). For such complex tasks, a problem designer can decompose it hierarchically. Then an expert can easily demonstrate it at different levels of implementation.

Another challenge with the IRL method is the design of feature spaces that capture the structure of the reward functions. Linear models for reward functions have been used in existing IRL algorithms. However, nonlinear models have recently been introduced \cite{Levine11}, \cite{Finn}, \cite{Metelli}. Exploring more general feature spaces for reward functions becomes necessary when expert intuition is insufficient for designing good features, including linear models. This problem raises concerns, such as in the robotics field \cite{Sermanet}.   

Regarding the first aspect of our problem, several works considered the decomposition of underlying reward functions for given expert demonstrations in RL and IRL problems (\cite{Henderson}, \cite{Choi}, \cite{Krishnan}). For hierarchical IRL problems, most of works focus on how to perform segmentation on demonstrations of complex tasks and find suitable reward functions. For the IRL problem in the options framework, option discovery should be first carried out as a segmentation process. Since our work focuses on hierarchical extensions of policy gradient based IRL algorithms, we assign options for each given domain instead of applying certain option discovery algorithms.

To simultaneously solve the reward construction problem while capturing the hierarchical structure, we propose a new method that applies the option framework presented by \cite{Sutton99} to the compatible reward inverse reinforcement learning (CR-IRL) \cite{Metelli}, a recent work on generating a feature space of rewards. Our method is called \textit{Option Compatible Reward Inverse Reinforcement Learning} (OCR-IRL). Previous works on the selection of proper reward functions for the IRL problem require design features that consider the environment of the problem. However, the CR-IRL algorithm directly provides a space of features from which compatible reward functions can be constructed.

The main contribution of our work comprises the following items.
\begin{itemize}
    \item New method of assigning reward functions for a hierarchical IRL problem is introduced. While handling the termination of each option, introducing parameters to termination and intra-option policy functions in the policy gradient framework allows us to choose better reward functions while reflecting the hierarchical structure of the task. 
    
    \item The recovered reward functions can be used to transfer knowledge across related tasks. Previous works such as \cite{Bacon17} have shown that the options framework provides benefits for transfer learning. Our method makes the knowledge transfer easier by converting the information contained in the options into a numerical reward value.
    
    \item It also shows better robustness to noise included in expert demonstrations than other algorithms without using a hierarchical learning framework. The noise robustness of our algorithm is enabled by general representation of reward functions compared to previous linear IRL algorithms.
\end{itemize}

There are differences in several aspects between our work and some of recent works \cite{Henderson}, \cite{Michini} and \cite{Krishnan} on segmentation of reward functions in IRL problems. Although both OptionGAN \cite{Henderson} and our work use policy gradient methods as a common grounding component, the former work adopts the generative adversarial approach to solve the IRL problem while we construct an explicit equation which defines reward features. \cite{Michini} uses Bayesian nonparametric mixture models to simultaneously partition the demonstration and learn associated reward functions. It has an advantage in the case with domains in which subgoals of each subtask are definite. For such domains, a successful segmentation simply defines task-wise reward functions. However, our work allows for indefiniteness of subgoals for which an assignment of rewards is not simple. \cite{Krishnan} focuses on segmentation using transitions defined as changes in local linearity about a kernel function. It assumes pre-designed features for reward functions. On the other hand, our method does not assume any pre-knowledge on feature spaces.

\section{Preliminaries}
\subsection{ Markov decision process} 
The Markov decision process comprises the state space, $\mathcal{S}$, the action space, $\mathcal{A}$, the transition function, $P : \mathcal{S} \times \mathcal{A} \rightarrow ( \mathcal{S} \rightarrow [0,1])$, and the reward function, $R: \mathcal{S} \times \mathcal{A} \rightarrow \R $. A policy is a probability distribution, $\pi : \mathcal{S} \times \mathcal{A} \rightarrow [0,1]$, over actions conditioned on the states. The value of a policy is defined as $V_{\pi} (s) = \mathbb{E}_{\pi} [ \sum_{t=0}^{\infty} \gamma^t R_{t+1} |S_0 =s ]$, and the action-value function is $Q_{\pi} (s, a) =\mathbb{E}_{\pi} [ \sum_{t=0}^{\infty} \gamma^t R_{t+1} |S_0 =s, A_0 = a ] $, where $\gamma \in[0, 1]$ is the discount factor. 
 
\subsection{Policy Gradients} 
Policy gradient methods \cite{Sutton00} aim to optimize a parametrized policy, $\pi_\theta$, via stochastic gradient ascent. In a discounted setting, the optimization of the expected $\gamma$-discounted return with respect to an initial state $s_0$, $\rho (\theta, s_0)=\mathbb{E}_{\pi_\theta} [ \sum_{t=0}^{\infty} \gamma^t R_{t+1} |S_0=s_0]$, is considered. It can be written as 
$$ 
    \rho(\theta, s_0) = \sum_s \mu_{\pi_\theta} (s |s_0)  \sum_a \pi_\theta (a|s) R(s,a)
$$
where $\mu_{\pi_\theta} (s|s_0) = \sum_{t=0}^{\infty} \gamma^t P (S_t=s| S_0=s_0, \pi_\theta )$.
The policy gradient theorem (\cite{Sutton00}) states:
$$ 
    \nabla_\theta \rho(\theta, s_0 ) = \sum_{s,a} \mu_{\pi_\theta} (s,a |s_0) \nabla_\theta \log\pi_\theta (a|s) Q_{\pi_\theta} (s,a),
$$
where $\mu_{\pi_\theta} (s,a|s_0) = \mu_{\pi_\theta} (s|s_0) \pi_\theta (a|s)$.
 
\subsection{Compatible Reward Inverse Reinforcement Learning}
Compatible reward inverse reinforcement learning\cite{Metelli} is an algorithm that generates a set of base functions spanning the subspace of reward functions that cause the policy gradient to vanish. As input, a parametric policy space, $\Pi_\Theta = \{ \pi_\theta : \theta \in \Theta \subset \R^k \}$, and a set of trajectories from the expert policy, $\pi^E$, are taken. It first builds the features, $\{ \phi_i \}$, of the action-value function, which cause the policy gradient to vanish. These features can be transformed into reward features, $\{ \psi_i \}$, via the Bellman equation (model-based case) or reward-shaping \cite{Ng99}(model-free). Then, a reward function that maximizes the expected return is chosen by enforcing a second-order optimality condition based on the policy Hessian \cite{Kakade}, \cite{Furmston}. 
 
\subsection{The options framework} 
We use the options framework\cite{Sutton99} which is a probability formulation for temporally extended actions. A Markovian option, $\omega \in \Omega $, is a triple $ ( I_{\omega}, \pi_{\omega}, \beta_{\omega}  ),$ where $ I_{\omega}$ is an initiation set, $ \pi_{\omega}$ is an intra-option policy, and $ \beta_{\omega}: \mathcal{S} \rightarrow [0, 1]$ is a termination function. Following \cite{Bacon17}, we consider the call-and-return option execution model in which the agent selects option $\omega$ according to the policy-over-options $\pi_\Omega(\omega|s)$ and follows the intra-option policy $\pi_\omega(a|s)$ until termination with probability $\beta_\omega(s)$. Let $\pi_{\omega, \theta}$ denote the intra-option policy of option $\omega$ parametrized by $\theta$ and $\beta_{\omega, \vartheta}$, the termination function of the same option parametrized by $\vartheta$.
 
%\subsection{Option-critic architecture} 
\cite{Bacon17} proposed a method of option discovery based on gradient descent applied to the expected discounted return, defined by 
$ 
    \rho (\Omega, \theta, \vartheta, s_0, \omega_0 ) 
    = E_{\Omega, \theta, \vartheta} \left[ \sum_{t=0}^{\infty} \gamma^t R_{t+1} | s_0, \omega_0  \right]. 
$
The objective function used here depends on policy-over-options and the parameters for intra-option policies and termination functions. Its gradient with respect to these parameters is taken through the following equations:
the option-value function can be written as 
$$ 
    Q_{\Omega} (s, \omega ) = \sum_{a} \pi_{\omega, \theta} (a|s) Q_U (s, \omega, a) 
$$
where
$$ 
    Q_U (s, \omega, a)= R(s,a) + \gamma \sum_{ s'} P (s'|s, a) U (\omega, s') 
$$
is the action-value function for the state-option pair, 
$$ 
    U (\omega, s')= (1- \beta_{\omega, \vartheta} (s') ) Q_{\Omega} (s', \omega) + \beta_{\omega, \vartheta} (s') V_{\Omega} (s') 
$$
is the option-value function upon arrival, and $V_{\Omega}$ is the value function over options.

\section{Generation of Q-features compatible with the optimal policy}
The first step to obtain a reward function as a solution for a given IRL problem is to generate Q-features (base functions of the action-value function space compatible with an expert policy) using the gradient of expected discounted returns. We assume that the parametrized expert intra-option policies, $\pi_{\omega, \theta}$, are differentiable with respect to $\theta$. By the intra-option policy gradient theorem \cite{Bacon17}, the gradient of the expected discounted return with respect to $\theta$ vanishes as in the following equation:
\begin{equation} \label{policy gradient}
 \nabla_\theta \rho =\sum_{s, \omega} \mu_{\Omega} (s, \omega |s_0, \omega_0 ) \sum_{a} \nabla_{\theta} \pi_{\omega, \theta} (a|s) 
Q_U (s, \omega, a) = 0
\end{equation} 
where $\mu_{\Omega} (s, \omega |s_0, \omega_0 )=\sum_{t=0}^\infty \gamma^t P(S_t=s, \omega_t=\omega|s_0,\omega_0)$ is the occupancy measure of state-option pairs. 

The first-order optimality condition, $\nabla_\theta \rho = 0 $, gives a defining equation for Q-features compatible with the optimal policy. It is convenient to define a subspace of such compatible Q-features in the Hilbert space of functions on $\Omega \times \mathcal{S} \times \mathcal{A}$. 
We define the inner product:
$$ 
    <f, g> := \sum_{\omega, s, a} f(\omega, s, a)  \mu_{\Omega} (s, \omega |s_0, \omega_0 ) \pi_{\omega, \theta} (a|s) g(\omega, s, a). 
$$
Consider the subspace, $G_{\pi} = \{ \nabla_{\theta} \log \pi_{\omega, \theta} \alpha: \alpha \in \R^k \}$, of the Hilbert space of functions on $\Omega \times \mathcal{S} \times \mathcal{A}$ with the inner product defined above. Then, the space of Q-features can be represented by the orthogonal complement, $ G_{\pi}^{\perp}$ of $G_\pi$.

Parametrization of terminations is expected to allow us to have more finely tuned option-wise reward functions in IRL problems. We can impose an additional optimality condition on the expected discounted return with respect to parameters of the termination function. Let $$\hat{\rho} (\Omega, \theta, \vartheta, s_1, \omega_0 )=E_{\Omega, \theta, \vartheta} [\sum_{t=0}^{\infty} \gamma^t R_{t+1} |\omega_0, s_1 ] $$
be the expected discounted return with initial condition $(s_1, \omega_0).$ By the termination gradient theorem \cite{Bacon17}, one has
\begin{equation} \label{termination gradient1}
 \nabla_{\vartheta} \hat{\rho}= - \sum_{s', \omega} \mu_{\Omega} (s', \omega | s_1, \omega_0 ) \nabla_{\vartheta} \beta_{\omega, \vartheta} (s') A_{\Omega} (s', \omega) \end{equation}
where $ A_{\Omega} $ is the advantage function over options $A_{\Omega} (s', \omega) = Q_{\Omega} (s', \omega) - V_{\Omega} (s') $.

The vanishing equation (\ref{termination gradient1}) gives a constraint on the space of the $Q$-feature, $ \hat{G}_{\pi}^{\perp}$. For simplicity, set $\mu_{1, \Omega} (s', \omega) =\mu_{\Omega} (s', \omega | s_1, \omega_0)$. The constraint equation for $ \hat{G}_{\pi}^{\perp}$ is given by
\begin{multline} \label{termination gradient}
 \sum_{\omega, s' }  \nabla_{\vartheta} \beta_{\omega, \vartheta} (s') \mu_{1, \Omega} (s', \omega  ) ( Q_{\Omega} (s', \omega )\\
 -\sum_{\omega'} \pi_{\Omega} (\omega'|s')  Q_{\Omega} (s', \omega' ) ) =0
\end{multline}
where $$ Q_{\Omega} (s, \omega ) = \sum_{a} \pi_{\omega, \theta} (a|s) Q_U (s, \omega, a). $$
Thus, we can combine two linear equations (\ref{policy gradient}), (\ref{termination gradient}) for $Q_U$ to define the space of $Q$-features.

\section{Reward function from Q-functions}

If two reward functions can produces the same optimal policy, then they satisfy the following(\cite{Ng99}): 
$$ 
    R' (s, a) = R(s, a) + \gamma \sum_{s'} P (s' |s, a) \chi(s') -\chi (s)
$$
for some state-dependent potential function $\chi$. This is called reward shaping. If we take $\chi = V$, then 
$$ 
    R' (s, a) = Q(s, a)- V (s) = Q(s,a) - \sum_{a'} \pi(a'|s) Q(s, a') 
$$

Because the $Q$-value function depends on the option in the options framework, the potential function, $\chi$, also depends on the option. We thus need to consider reward-shaping with regards to the intra-option policy, $\pi_{\omega}$. Then, the reward functions also need to be defined in the intra-option sense. This viewpoint is essential to our work and is similar to the approach taken in \cite{Henderson}, in which $R_\omega$, the reward option, was introduced corresponding to the intra-option policy, $\pi_\omega$. Reward functions, $R_{\omega}, R'_{\omega}$, sharing the same intra-option policy, $\pi_\omega$, satisfy $$ R'_\omega (s, a) = R_\omega (s, a) + \gamma \sum_{s'} P (s' |s, a) \chi(s',\omega) -\chi (s,\omega). $$
If we take $\chi (s,\omega)=U(\omega, s) $, then 
\begin{align*}
   R'_{\omega} (s,a) 
   &= R_{\omega} (s, a) + \gamma \sum_{s'} P (s' | s,a) U (\omega, s') - U (\omega, s) \\
   &= Q_U  (s,\omega, a) - [ (1-\beta (s) ) Q_{\Omega}(\omega, s)  + \beta (s) V_{\Omega} (s)] \\
   &= Q_U  (s,\omega, a) - \sum_{a'} \pi_{\omega} (a |s) Q_U  (s,\omega, a) +\beta (s) A_{\Omega} (s,\omega)
\end{align*}
This provides us with a way to generate reward functions from $Q$-features in the options framework.

\section{Reward selection via the second-order optimality condition}

Among the linear combinations of reward features constructed in the previous section, selecting a linear combination that maximizes $\rho (\theta)$ and $\hat{\rho} (\vartheta)$ is required. For the purpose of optimization, we use the second-order optimality condition based on the Hessian of $\rho (\theta)$ and $\hat{\rho} (\vartheta)$.

Consider a trajectory, $\tau = ((s_0, \omega_0, a_0, b_0), \cdots, (s_{T-1}, \omega_{T-1}, a_{T-1}, b_{T-1}))$, with termination indicator $b_t$ and terminal state $s_T$. The termination indicator, $b_t$, is 1 if a previous option terminates at step $t$, otherwise 0. The probability density of trajectory $\tau$ is given by 
\begin{multline*}
    \mathbb{P}_{\theta, \vartheta}(\tau) = p_0(s_0) \delta_{b_0 = 1} \pi_\Omega(\omega_0 | s_0) \prod_{t=1}^{T-1} \mathbb{P} (b_t, \omega_t | \omega_{t-1}, s_t) \\
    \prod_{ t=0}^{T-1} \pi_{\omega_t} (a_t | s_t) p(s_{t+1} | s_t, a_t), 
\end{multline*}
where
\begin{align*}
    \mathbb{P}(b_t = 1, \omega_t | \omega_{t-1}, s_t) &= \beta_{\omega_{t-1}} (s_t) \pi_\Omega(\omega_t | s_t) \\
    \mathbb{P}(b_t = 0, \omega_t | \omega_{t-1}, s_t) &= (1 - \beta_{\omega_{t-1}} (s_t)) \delta_{\omega_t = \omega_{t-1}}.
\end{align*}
We denote the space of all possible trajectories by $\mathbb{T}$ and the $\gamma$-discounted trajectory reward by $R(\tau) = \sum_{t=0}^{T(\tau)-1} \gamma^t R(s_{\tau, t}, a_{\tau, t})$. Then, the objective function can be rewritten as
$$
    \rho(\Omega, \theta, \vartheta, s_0, \omega_0) 
    = E[ \sum_{t=0}^\infty \gamma^t R_{t+1} | s_0, \omega_0 ] 
    = \int_{\mathbb{T}} \mathbb{P}_{\theta, \vartheta}(\tau) R(\tau) d\tau.    
$$
Its gradient and Hessian with respect to $\theta$ can be expressed as
$$
    \nabla_\theta \rho 
    = \int_\mathbb{T} \mathbb{P}_{\theta, \vartheta}(\tau) \nabla_\theta \log \mathbb{P}_{\theta, \vartheta}(\tau) R(\tau) d\tau
$$
and
\begin{multline*}
    \mathcal{H}_{\theta} \rho
    = \int_\mathbb{T} \mathbb{P}_{\theta, \vartheta}(\tau) (\nabla_\theta \log \mathbb{P}_{\theta, \vartheta}(\tau) \nabla_\theta \log \mathbb{P}_{\theta, \vartheta}(\tau)^T \\
    + \mathcal{H}_\theta \log \mathbb{P}_{\theta, \vartheta}(\tau)) R(\tau) d\tau.
\end{multline*}
The second objective function can be written as
$$
    \hat{\rho} = E[ \sum_{t=0}^\infty \gamma^t R_{t+1} | \omega_0, s_1 ] 
    = \int_{\hat{\mathbb{T}}} \hat{\mathbb{P}}_{\theta, \vartheta}(\hat{\tau}) R(\hat{\tau}) d\hat{\tau},     
$$
where $\hat{\tau}$ is a trajectory beginning with $(\omega_0, s_1)$ with the probability distribution 
\begin{multline*}
    \hat{\mathbb{P}}_{\theta, \vartheta}(\hat{\tau}) = p_0(\omega_0, s_1) \prod_{t=1}^{T-1} \mathbb{P} (b_t, \omega_t | \omega_{t-1}, s_t) \\
    \prod_{ t=1}^{T-1} \pi_{\omega_t} (a_t | s_t) p(s_{t+1} | s_t, a_t).  
\end{multline*}
Then, its Hessian can be written as
\begin{multline*}
    \mathcal{H}_{\vartheta} \hat{\rho}
    = \int_{\hat{\mathbb{T}}} \hat{\mathbb{P}}_{\theta, \vartheta}(\hat{\tau}) (\nabla_\vartheta \log \hat{\mathbb{P}}_{\theta, \vartheta}(\hat{\tau}) \nabla_\vartheta \log \hat{\mathbb{P}}_{\theta, \vartheta}(\hat{\tau})^T \\+ \mathcal{H}_\vartheta \log \hat{\mathbb{P}}_{\theta, \vartheta}(\hat{\tau})) R(\hat{\tau}) d\hat{\tau}.
\end{multline*}

Let $\{ \psi_{\omega, i} \}$ be the reward features constructed from the previous section. Rewrite each Hessian as $\mathcal{H}_{\theta} (\rho) = \sum_i w_i \mathcal{H}_{\theta} \rho_i,$ where $\rho_i$ is the expected return with respect to $  \mathbb{P}_{\theta, \vartheta}$ for the reward function, $\psi_i$, and as $\mathcal{H}_{\vartheta} (\hat{\rho}) = \sum_i w_i \mathcal{H}_{\vartheta} \hat{\rho}_i,$ where $\hat{\rho}_i$ is the expected return with respect to $  \hat{\mathbb{P}}_{\theta, \vartheta}$ for the reward function, $\psi_i$.
 Set $tr_{\theta, i} = tr (\mathcal{H}_{\theta} (\rho_i) )$ and $tr_{\vartheta, i} = tr (\mathcal{H}_{\vartheta} (\hat{\rho}_i) )$ for $i=1, \cdots, p$.

To choose the reward features that achieve local maxima of the objective functions, we only need to observe whether each Hessian matrix is negative definite. This can be done by imposing the constraint that the maximum eigenvalue of the Hessian is negative. In practice, however, the strict negative definiteness is rarely met. Analysis for this result is presented in \cite{Metelli}. As alternative, we determine the reward weight, $\mathbf{w}$, for the reward function, $R_\omega = \sum_{i=1}^p w_{i} \psi_{\omega, i}$, which yields a negative semi-definite Hessian with a minimal trace. 
Also, to relieve a computational burden, we exploit a heuristic method suggested by \cite{Metelli}: we only choose reward features having negative definite Hessians, compute the trace of each Hessian, and collect them in the vectors $\mathbf{tr}_{\theta} = (tr_{\theta, i})$ and $\mathbf{tr}_{\vartheta} = (tr_{\vartheta, i})$. We determine $\textbf{w}$ by solving 
$$ 
    \min_{\textbf{w} } \textbf{w}^{T} \textbf{tr}_{\theta}, \quad \mbox{ and} \quad  \min_{\textbf{w} } \textbf{w}^{T} \textbf{tr}_{\vartheta} \qquad \mbox{s. t. }  \qquad \| \mathbf{w}\|^2_2 =1.
$$
Typically, multi-objective optimization problems have no single solutions that optimize all objective functions simultaneously. One well-known approach to tackling this problem is a linear scalarization. Thus, we consider the following single-objective problem:
$$
    \min_{\textbf{w}} \lambda_\theta \textbf{w}^{T} \textbf{tr}_{\theta} + \lambda_\vartheta \textbf{w}^{T} \textbf{tr}_{\vartheta}  \qquad
  \mbox{s. t. }  \qquad \| \mathbf{w}\|^2_2 =1
$$
with positive weights $\lambda_\theta$ and $\lambda_\vartheta$. A closed-form solution is computed as
 $\mathbf{w} = - (\lambda_\theta \textbf{w}^{T} \textbf{tr}_{\theta} + \lambda_\vartheta \textbf{w}^{T} \textbf{tr}_{\vartheta}) / \|\lambda_\theta\textbf{w}^{T} \textbf{tr}_{\theta} + \lambda_\vartheta \textbf{w}^{T} \textbf{tr}_{\vartheta}\|$.
With a different choice of scalarization weights, $\lambda_\theta$ and $\lambda_\vartheta$, different reward functions can be produced. It is natural to set $\lambda_\theta = 1/\|\textbf{tr}_{\theta} \|$ and $\lambda_\vartheta = 1/\|\textbf{tr}_{\vartheta} \|$ because the gap between the magnitudes of two trace vectors can be large in practice. 
Here, we can guarantee the obtained solution is Pareto optimal.

\section{Algorithm}
We summarize our algorithm of solving the IRL problem in the options framework as follows:
\begin{algorithm}
    \scriptsize
    \caption{Option Compatible Reward IRL}
    \begin{algorithmic}[1]
        \renewcommand{\algorithmicrequire}{\textbf{Input:}}
        \renewcommand{\algorithmicensure}{\textbf{Output:}}
        
        \REQUIRE (1) Expert's demo-trajectories 
        $D= \cup_{i=1}^N \{ (\omega_{\tau_i, 1}, s_{\tau_i, 0}, a_{\tau_i, 0} ), (\omega_{\tau_i, 1}, s_{\tau_i, 1}, a_{\tau_i, 1} ), \allowbreak\cdots,  (\omega_{\tau_i, T(\tau_i)}, s_{\tau_i,T(\tau_i) }, a_{\tau_i, T(\tau_i} ) \}$, 
        (2) Option $\Omega_{\theta, \vartheta}=\{ \omega \}$, for which expert's policies, $\pi_{\omega, \theta}$, and termination functions, $\beta_{\omega, \vartheta}$, are parametrized, and a policy $\pi_{\Omega}$ over options.
        \ENSURE Reward function $R_{\omega} (s, a). $
        
        \textbf{Phase 1}
        \STATE Estimate $\mu =\mu_{\Omega} (s, \omega | s_0, \omega_0 )  \pi_{\omega, \theta} (a|s) $ for the visited state-option-action triples from the trajectories.
        \STATE Get the matrices $ D_{\Omega}=\mbox{ diag }( \mu ) $ and $ \nabla_{\theta} \log \pi_{\omega} (a|s) $.
        \STATE Find the basis for the null space of $\nabla_{\theta} \log \pi_{\omega}^T   D_{\Omega}$ (e.g. using singular value decomposition).
        \STATE Estimate $\mu_1 = \mu_{\Omega} (s, \omega | s_1, \omega_0) $  for the visited option-state pairs.
        \STATE Get the matrices $ \mbox{diag} (\mu_1)$, $\nabla_{\vartheta} \beta $, $\Pi_{\Omega}$, and $\Pi$. 
        \STATE  Find the basis for the null space of $ \nabla_{\vartheta} \beta^T \mbox{diag}( \mu_1) (I - \Pi_\Omega ) \Pi $.
        \STATE Find the intersection, $\Phi$, of two null spaces.
        
        \textbf{Phase 2}
        \STATE Get the set of advantage functions using $A=(I- \Pi_{\Omega}) \Pi \Phi$. 
        \STATE Get the set of reward functions by applying reward shaping $ \Psi = (I - \Pi) \Phi + \beta A $. 
        %\STATE Apply singular value decomposition to orthogonalize $\Psi$.
        \STATE Estimate the Hessian, $\hat{\mathcal{H}}_{\theta} \rho_i$ and $\hat{\mathcal{H}}_{\vartheta} \hat{\rho}_i$, for each reward feature, $\psi_i$, $i=1, \ldots p$%, using equations:

        \STATE Discard the reward feature having an indefinite Hessian; switch sign for those having positive semi-definite Hessian; and compute 
$tr_{\theta, i} = tr (\hat{\mathcal{H}}_{\theta} (\rho_i) )$ and $tr_{\vartheta, i} = tr (\hat{\mathcal{H}}_{\vartheta} (\hat{\rho}_i) )$ for $i=1, \cdots, p$
        \STATE Reward function $R_\omega = \Psi \mathbf{w}_\omega$, $\mathbf{w}_\omega = -(1/\sqrt{2})(  \textbf{tr}_\theta / \| \textbf{tr}_\theta \|  
+ \textbf{tr}_\vartheta / \| \textbf{tr}_\vartheta \| ) $

    \end{algorithmic}
\end{algorithm}

Our algorithm consists of three phases. In the first phase, we obtain basis for Q-features space by solving linear equations. Linear equations consist of two parts. The first part is defined by the gradient of logarithmic policy  and the second part is defined by the gradient of option termination. The matrices $\Pi_\Omega$ and $\Pi$ are introduced to carry out computation for the second part. The matrix $\Pi_\Omega$ is the row repetition of policy over option, $\pi_\Omega$, on visited option and state pair. The matrix $\Pi$ is a block diagonal where each entry is intra-option policy over visited state and action pair for each option. 

In the second phase, we obtain basis for reward-features using reward shaping for option. In the last phase, we select the definite reward by applying Hessian test to two objective functions.

Our algorithm can be naturally extended to continuous states and action spaces. In the continuous domains we use a $k$-nearest neighbors method to extend recovered reward functions to non-visited state-action pairs. Additional penalization terms can be included. Details about implementation are presented in section 7.2. 

\section{Experiment}

\subsection{Transfer Learning}

\begin{figure*}[!t]
\centering
\includegraphics[width=\textwidth]{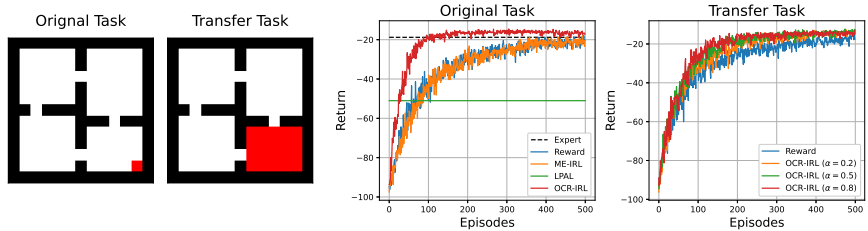}
\caption{Left two figures are the grid worlds in our setting, in which red color indicates goal locations. Right two figures are the plotting of average return of Fourrooms domain as a function of the number of episodes used in training for original task and transfer task, respectively.}
\label{four_return}
\end{figure*}

We test our method in a navigation task in the four-rooms domain suggested in \cite{Sutton99}. Our goal is to verify that our method can transfer knowledge between different environments but with similar tasks. 

First, a reward function is recovered by applying our method to the set of options which is learned in an original environment. The recovered reward function will be used to train in modified environments. To be specific, an initial goal state is located in the lower right corner, whereas the goal moves to a random location in the lower right room in the modified environments. Left two gridworlds in figure \ref{four_return} describe each environment in our setting in which red cells represent possible goal locations to be reached. The initial states are randomly chosen in the upper left room in the both cases. The possible actions are movements in four directions, which can be failed with probability $1/3$, in which case the agent takes random actions. The default reward is $-1$ for each step, and 0 when reaching to the goal cell. We evaluate our method based on options discovered by \cite{Bacon17}. To be specific, the policy over options and intra-option policies are parametrized as Boltzmann policies, and the terminations as sigmoid functions. 

For comparison, we give weights to the option-wise reward function, $R_\omega(s,a)$, based on the policy over options:
$$
    R(s,a) = \sum_{\omega \in \Omega} \pi_\Omega(\omega|s) R_\omega(s,a).
$$
It is easy to compare against other IRL algorithms by combining the rewards assigned to each option while the modified reward $R(s,a)$ maintains the nature of each task. We first evaluate OCR-IRL against maximum entropy IRL (ME-IRL) \cite{Ziebart} and linear programming apprenticeship learning (LPAL) \cite{Abbeel} in the original task. In this case a tabular representation for state is used for a reward feature in ME-IRL and LPAL. Figure \ref{four_return} show the results of training a Boltzmann policy using SARSA, coped with the default reward function and the recovered reward functions by each algorithms. Each result is averaged over 20 repetitions, using 50 expert demonstrations which are generated by the option discovered. We see that the reward obtained by OCR-IRL converges faster to the optimal policy than does the default reward function and ME-IRL. Despite that the input demonstrations are near-optimal, the reward recovered by our method guarantees learning the optimal policy, as shown in figure \ref{four_return}.

On the other hand, the rightmost plot in figure \ref{four_return} shows that our reward function can be used to accelerate learning in the transfer tasks. In order to incorporate our reward function to the default reward, we simply use a weighted sum of two rewards with different weights: 
$$
    R(s,a) = (1-\alpha) R_{default}(s,a) + \alpha R_{OCR-IRL}(s,a).
$$
The larger the value of $\alpha$, the more information, including the hierarchical structure of options, learned in the original domain can be delivered. We observe that the case for $\alpha=0.8$ outperforms the other cases. The reward recovered by ME-IRL has no effect on transfer.

 \subsection{Car on the Hill}
 
\begin{figure*}[!t]
\centering
\includegraphics[width=\textwidth]{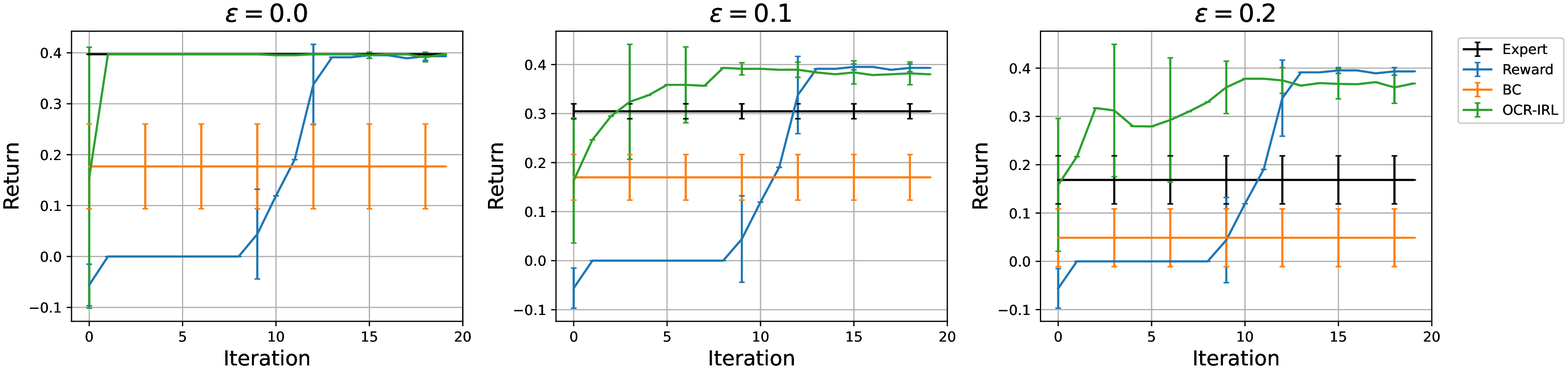}
\caption{Average return of Car-on-the-Hill domain as a function of the number of FQI iterations.}
\label{car_return}
\end{figure*}

\begin{figure*}[!t]
\centering
\includegraphics[width=\textwidth]{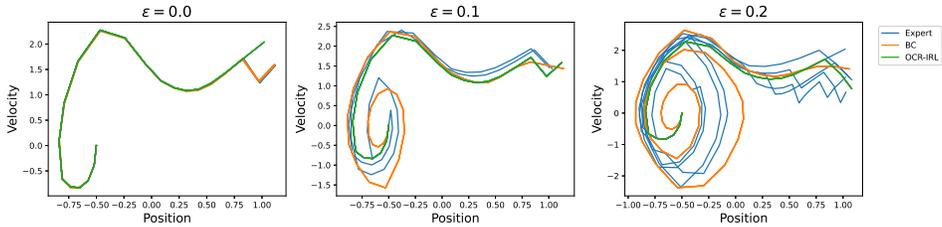}
\caption{Trajectories of the expert’s policy, the BC policy, and the policy computed via FQI with the reward recovered by OCR-IRL.}
\label{car_traj}
\end{figure*}

 We test OCR-IRL in the continuous Car-on-the-Hill domain \cite{Ernst}. A car traveling on a hill is required to reach the top of the hill. Here, the shape of the hill is given by the function, $Hill(p)$:
 $$
    Hill(p) = \left\{ \begin{array}{c}
         p^2 + p \quad\text{if}\ p<0 \\
         \frac{p}{\sqrt{1 + 5p^2}} \quad\text{if}\ p\geq0.
    \end{array}\right.
 $$ 
 The state space is continuous with dimension two: position $p$ and velocity $v$ of the car with $p \in [-1, 1]$ and $v \in [-3, 3]$. The action $a \in [-4, 4]$ acts on the car’s acceleration. The reward function, $R(p, v, a)$, is defined as:
 $$
    R(p_t, v_t, a_t) = \left\{ \begin{array}{l}
         -1 \quad\text{if}\ p_{t+1} < -1 \ \text{or}\ |v_{t+1}|>3  \\
         1 \quad\text{if}\ p_{t+1} > 1 \ \text{and}\ |v_{t+1}|\leq 3 \\
         0 \quad\text{otherwise}.
    \end{array}\right.
 $$
 The discount factor, $\gamma$, is 0.95, and the initial state is $p_0= -0.5$ with $v_0=0$.
 
 Because the car engine is not strong enough, simply accelerating up the slope cannot make it to the desired goal. The entire task can be divided into two subtasks: reaching enough speed at the bottom of the valley to leverage potential energy (subgoal 1), and driving to the top (subgoal 2). To evaluate our algorithm, we introduce hand-crafted options:
 $$
 \left\{ \begin{array}{l}
      I_\omega: \quad \text{the state space}\ S \\
      \pi_\omega: \quad \text{the policy for subgoal}\ \omega  \\
      \beta_\omega: \quad 1\ \text{if the agent achieves the subgoal}; 0\ \text{otherwise} 
 \end{array}\right.
 $$
 for $\omega \in \{1, 2\}$. Intra-option policy $\pi_\omega$ is defined by approximating the deterministic intra-option policies, $\pi_{\omega, FQI}$, via the fitted-Q iteration (FQI) \cite{Ernst} with the two corresponding small MDPs. We consider noisy intra-option policies in which a random action is selected with probability $\epsilon$:
$$
    \pi_\omega(a|s) = (1-\epsilon) \pi_{\omega, FQI}(a|s) + \epsilon \pi_{random}(a|s)
$$
for each option, $\omega$. Each intra-option policy is parametrized as Gaussian policy $\pi_{\theta,\omega}(a|s) \sim \mathcal{N}(y_{\theta,\omega}(s), \sigma^2)$, where $\sigma^2$ is fixed to be 0.01, and $y_{\theta,\omega}(s)$ is obtained using radial basis functions:
$$
    y_{\theta,\omega}(s) = \sum_{k=1}^N \theta_{\omega, k} e^{-\delta \|s-s_k\|^2},
$$
with uniform grids, $s_k$, in the state space. The parameter, $\theta_\omega$, is estimated using 20 expert trajectories for each option. Termination probability, $\beta_{\vartheta, \omega}$, is parametrized as a sigmoid function. 

For comparison, the task-wise reward function, $R_\omega(s,a)$, is merged into one reward, $R(s,a)$, by omitting the option term. This modification is possible, because the policy-over-options is deterministic in our setting. The merged reward function, $R(s,a)$, can be compared with other reward functions using a non-hierarchical RL algorithm. 

We extend the recovered reward function to non-visited state-action pairs using a kernel $k$-nearest neighbors (KNN) regression with a Gaussian kernel:
$$
    \hat{R}(s,a) = \frac{\sum_{(s',a')\in\text{KNN}((s,a),k,\mathcal{D})}\mathcal{K}((s,a),(s',a'))R(s',a')}{\sum_{(s',a')\in\text{KNN}((s,a),k,\mathcal{D})}\mathcal{K}((s,a),(s',a'))}
$$
where $\text{KNN}((s,a),k,\mathcal{D})$ is the set of the $k$ nearest state-action pairs in the demonstrations, $\mathcal{D}$, and $\mathcal{K}$ is a Gaussian kernel over $S\times A$:
$$
    \mathcal{K}((s,a),(s',a'))=\exp\left(-\frac{1}{2\sigma_S^2}\|s-s'\|^2 - \frac{1}{2\sigma_A^2}\|a-a'\|^2\right). 
$$
These reward extension is based on \cite{Metelli}.

The recovered rewards are obtained from expert demonstrations with different levels of noise, $\epsilon$. We repeated the evaluation over 10 runs. As shown in Figure \ref{car_return}, FQI with the reward function outperforms the original reward in terms of convergence speed, regardless of noise level. When $\epsilon=0$, OCR-IRL converges to the optimal policy in only one iteration. As the noise level $\epsilon$ increases, BC yields worse performance, whereas OCR-IRL is still robust to noise. 

Figure \ref{car_traj} displays the trajectories of the expert's policy, the BC policy, and the policy computed via FQI with the reward recovered by OCR-IRL. When $\epsilon = 0$, trajectories are almost overlapping. When $\epsilon$ increases, BC requires more steps to reach to the terminal state, and some cannot finish the task properly. On the other hand, we see that our reward function can recover the optimal policy, even if expert demonstrations are not close to optimal.

\section{Conclusion}
We developed a model-free IRL algorithm for hierarchical tasks modeled in the options framework. Our algorithm, OCR-IRL, extracts reward features using first-order optimality conditions based on the gradient for intra-option policies and termination functions. Then, it constructs option-wise reward functions from the extracted reward spaces using a second-order optimality condition. The recovered reward functions explain the expert's behavior and the underlying hierarchical structure. 

Most IRL algorithms require hand-crafted reward features, which are crucial to the quality of recovered reward functions. Our algorithm directly builds the approximate space of the reward function from expert demonstrations. Additionally, unlike other IRL methods, our algorithm does not require solving a forward problem as an inner step.

Some heuristic methods were used to solve the multi-objective optimization problem in the reward selection step. We used linear scalarization to change the problem to a single-objective optimization problem, empirically finding that this approach resulted in good performances. Generally, depending on the type of option used, one of parameters of intra-option policies or termination functions could be more sensitive than the other. Therefore, the choice of weights can make a difference in the final performance. 

Our algorithm was validated in several classical benchmark domains, but to apply it to real-world problems, we need to experiment with more complex environments. More sophisticated options should be used to better explain the complex nature of a hierarchical task, making experiment extensions easier.

\section*{Acknowledgments}
This work was supported in part by the National Research Foundation of Korea (NRF) grant funded by the Korea Government (MSIT) under Grant 2017R1E1A1A03070105, and in part by the Institute for the Information and Communications Technology Promotion (IITP) grant funded by the Korea Government (MSIP) [Artificial Intelligence Graduate School Program (POSTECH) under Grant 2019-0-01906 and the Information Technology Research Center (ITRC) Support Program under Grant IITP-2018-0-01441].

\bibliographystyle{plain}
\bibliography{refs.bib}

\end{document}